\documentclass[12pt]{article}
\usepackage{float}

\usepackage[utf8]{inputenc}
\usepackage[left=1.2in,right=1.2in,top=1.2in,bottom=1.2in]{geometry}

\usepackage{amsmath, amssymb, amsfonts, amsthm, mathtools, nccmath}

\usepackage{xurl}
\usepackage[
  colorlinks=true,
  linkcolor=black,
  citecolor=black,
  urlcolor=blue
]{hyperref}

\usepackage{graphicx}
\graphicspath{{./figures/}}
\usepackage{booktabs}
\usepackage{multirow}
\usepackage{subcaption}
\usepackage{caption}

\usepackage{enumitem}
\usepackage{multicol}
\usepackage{setspace}

\usepackage{float}
\usepackage{natbib}
\usepackage{url}

\usepackage{tikz}
\usetikzlibrary{matrix,backgrounds}
\usepackage{pgfplots}
\pgfplotsset{compat=1.16}

\usepackage{listings}
\usepackage{verbatim}
\usepackage{appendix}
\usepackage{authblk}
\usepackage{lineno}
\usepackage{cancel}
\usepackage{arydshln}
\usepackage[normalem]{ulem}
\usepackage{pgfkeys}

\title{Integrated Marketing Attribution: A Bayesian Framework for Privacy-Safe Granular Measurement Anchored in MMM }
\author{Meghana R. Bhat}
\author{Ankit Umare}

\author{Utsav Aggarwal}
\author{Richard Vecsler}
\author{Arunkumar Mani}
\author{Karthik Nair}
\author{Chandhu Nair}
\affil{Lowe's Companies, Inc.} 
\date{}

\begin{document}

\maketitle

\begin{abstract}
Retail marketing measurement increasingly requires granular campaign-level insights without relying on user-level tracking. However, the two dominant approaches, Marketing Mix Modeling (MMM) and Multi-Touch Attribution (MTA), often produce fragmented insights. MMM is privacy-safe and robust for channel-level planning but is too coarse for campaign optimization, while MTA provides granular attribution but has become less reliable under increasing privacy restrictions. We propose \textbf{Integrated Marketing Attribution (IMA)}, a unified framework that combines MMM with channel-specific Bayesian attribution models to derive campaign-level effects from aggregated data. By leveraging MMM-informed priors, IMA delivers granular, privacy-safe attribution while preserving consistency with MMM.

\end{abstract}

\section{Introduction}
Modern marketing measurement is increasingly defined by a tradeoff between robustness and granularity. Retailers and brands need measurement systems that can guide strategic budget allocation while also supporting tactical campaign optimization. Two approaches have traditionally dominated this space: MMM and MTA. While both provide valuable insights, they operate at different levels of granularity and rely on distinct data sources and assumptions, often resulting in fragmented measurement and misalignment between strategic planning and tactical execution. 

MMM estimates the impact of marketing activities using aggregated data over long time horizons, making it a robust and privacy-safe approach for measuring channel-level effectiveness across both digital and offline media. MMM is therefore widely used to inform longer-horizon budget planning and channel-level investment decisions. However, it is poorly suited for evaluating individual campaigns. At the campaign-level granularity, model dimensionality increases substantially, correlations across campaigns intensify, and the number of observations becomes limited, leading to unstable parameter estimates and overfitting. As a result, MMM alone cannot reliably support short-term decision-making. 

MTA  was introduced to address these limitations by assigning conversion credit to individual digital touchpoints along a customer journey. By leveraging user-level data, MTA provides quick feedback for campaign optimization. However, the ongoing shift toward a privacy-centric digital ecosystem, including the deprecation of third-party cookies and restrictions on cross-site tracking, has significantly reduced the reliability and scalability of MTA. As customer journeys become increasingly incomplete or unobservable, MTA models suffer from biased attribution and operational fragility. 

These limitations reveal a critical gap in modern marketing measurement: the absence of a privacy-safe framework that delivers granular, actionable insights while remaining consistent with MMM outcomes. To address this gap, we propose Integrated Marketing Attribution (IMA), a unified modeling framework that delivers granular attribution using aggregated data and is tightly integrated with MMM. IMA decomposes channel-level incremental contributions estimated by MMM into campaign-level effects using channel-specific attribution models. By adopting MMM-informed Bayesian priors, IMA mitigates high dimensionality and multicollinearity while preserving consistency with MMM outputs.

The contributions of this paper are threefold: (1) we introduce IMA as a unified and privacy-safe attribution framework; (2) we propose a channel-specific Bayesian modeling approach guided by MMM priors; and (3) we demonstrate the effectiveness of IMA using real-world marketing data.

\section{Related Work}

MTA has been widely used for granular digital measurement. However, the loss of third-party cookies and restrictions on cross-site tracking reduce visibility into complete customer journeys, rendering user-level attribution models increasingly infeasible \citep{churchill2025modeling}. As a result, MTA suffers from incomplete data, biased attribution, and reduced reliability, limiting its usefulness in a privacy-centric environment \citep{greenfield2022cookies}.

In response, alternative approaches to marketing measurement have gained prominence. MMM has re-emerged as a scalable and privacy-safe approach based on aggregated data, supported by recent advances in Bayesian MMM and related methodologies \citep{jin2017bayesian, chan2017challenges}. At the same time, \citet{gordon2019comparison} show that incrementality-based methods based on randomized experiments provide a principled framework for measuring causal effects. More recently, \citet{Google2022UMM} propose a unified measurement framework that integrates MMM, MTA, and experimentation within a Bayesian setting, emphasizing knowledge transfer via priors, modeling cross-channel interactions, and generating consistent probabilistic insights for improved decision-making. Related work also explores aggregate-data approaches motivated by privacy constraints \citep{churchill2025modeling}, combining elements of MMM and machine learning.

These developments are also reflected in industry perspectives. For example, \citet{forbes2023cookieless, forbes2022mmm} highlight MMM, incrementality experiments, and hybrid frameworks as practical alternatives to MTA, emphasizing trade-offs between granularity, scalability, and causal rigor, while \citet{skai2022cookieless} advocate for incrementality-based measurement to isolate true marketing impact under privacy constraints.

Several works position MMM as a central component of future measurement systems. \citet{mass2022mmm_future} argue that MMM will play a key role due to its reliance on aggregated data and privacy-safe design, highlighting advances such as increased granularity, faster refresh cycles, and Bayesian methods. \citet{lau2022amex_mmm} describe how American Express is adopting MMM as a privacy-safe foundation while extending it to support more granular, campaign-informed decisions through geo-level analysis and richer digital signals. However, neither work provides a reproducible, end-to-end methodology validated across diverse campaign types.

While incrementality-based approaches provide strong causal insights, they remain difficult to scale in practice, particularly for large retail organizations running thousands of campaigns across channels and geographies. Designing, executing, and maintaining experiments at this scale is often operationally and economically infeasible. Motivated by these constraints, we adopt a granular MMM-based approach in IMA. Unlike prior proposals that often lack end-to-end validation on real-world campaign data at scale, IMA is a concrete methodology validated using real-world marketing data.

\section{Integrated Marketing Attribution (IMA) Framework}

IMA is designed to decompose channel-level incremental contributions estimated by MMM into granular campaign-level effects. Guided by Bayesian priors derived from MMM, the framework estimates separate models for each channel, mitigating high dimensionality and cross-channel multicollinearity without relying on user-level tracking.

\begin{figure}[h]
    \centering
    \includegraphics[width=0.8\linewidth]{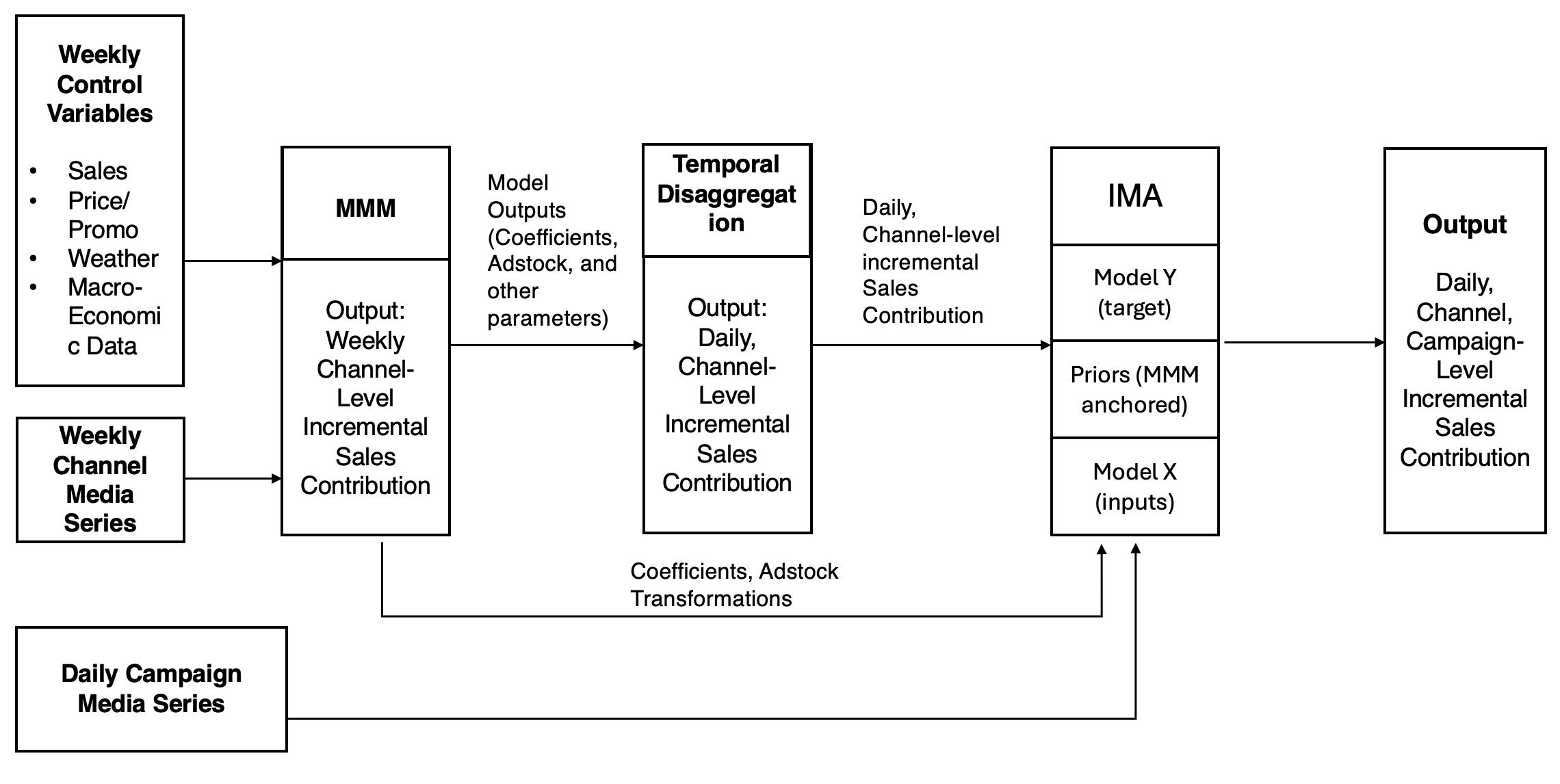}
    \caption{IMA Model Architecture}
    \label{fig:ima_model_arch}
\end{figure}

As illustrated in Figure~\ref{fig:ima_model_arch}, the IMA framework consists of three stages: (1) MMM estimation, (2) temporal disaggregation, and (3) channel-specific attribution modeling. Table~\ref{tab:ima_schema} summarizes the underlying data schema and modeling structure used throughout the framework. While MMM operates on weekly channel-level data, IMA produces daily campaign-level attribution by disaggregating channel-level incremental sales contributions using daily media series and channel-specific models. 

First, a weekly MMM model is estimated using aggregated sales, pricing, promotion, macroeconomic variables, and weekly channel-level media series. MMM produces two key outputs: (i) weekly channel-level incremental sales contributions and (ii) channel-level adstock parameters and coefficients. 

Second, the weekly channel-level incremental contributions are temporally disaggregated to daily resolution. Rather than applying a naïve proportional split, we leverage the adstock transformations learned by MMM. For each channel, the MMM-estimated adstock parameters are converted from weekly to daily scale and tuned where necessary to reflect daily signal dynamics. These daily adstock parameters are then applied to daily channel-level media data to generate adstock-transformed daily media series which are subsequently used to distribute weekly channel-level incremental sales into daily channel-level incremental contributions. This preserves the carryover structure learned by MMM while enabling daily-level modeling. 

Finally, IMA trains independent attribution models for each channel. 

\begin{table}[htbp]
\centering
\caption{Data schema and modeling structure used in the IMA framework}
\label{tab:ima_schema}

\begin{tabular}{ll}
\toprule
\textbf{Field} & \textbf{Description} \\
\midrule

Modeling window & 3 years historical data \\
MMM intervals & Weekly \\
IMA intervals & Daily \\
Channels & Search, Social, Display,... \\
Media series & Impressions, clicks, or spend \\
Outcome & Campaign-level incremental sales  \\
Evaluation & Channel-level aggregated IMA vs. MMM contributions  \\
\bottomrule
\end{tabular}

\end{table}

\subsection{Channel-Specific Attribution Models}

For each channel $c$, we estimate a separate Bayesian regression model. The target variable (Model $Y$) is the daily channel-level incremental sales contribution obtained from temporal disaggregation. The input variables (Model $X$) consist of daily campaign-level media series within channel $c$, such as spend, impressions, and clicks. The estimated campaign-level contributions are computed using the fitted coefficients from the channel-specific IMA model and the corresponding adstock-transformed campaign-level media series. Channel-level coefficients estimated by MMM serve as informative priors for the corresponding channel-specific model, guiding campaign-level estimates while reducing overfitting when campaign-level data are sparse or highly correlated. This formulation ensures that the estimated campaign-level contributions aggregate to the daily channel-level incremental sales anchored by MMM.

\subsection{IMA Model Specification}

We do not elaborate on the MMM specification itself, as MMM is a well-established modeling framework and the focus of this paper is the downstream attribution layer that uses MMM outputs as anchors. 
\paragraph{Adstock Transformation.}
Media effects on sales may lag behind the original exposure and persist over several periods. Following \citet{jin2017bayesian} and \citet{he2020mmm}, we model this carry-over effect using an adstock transformation with parameters $L$ (length of the media effect window), $P$ (peak/delay of the effect), and $D \in (0, 1)$ (decay/retention rate).

For a campaign-level media input series $\{x_t\}$, where $x_t$ denotes the raw media exposure at time $t$, the adstock-transformed value $x_t^*$ is computed as a normalized weighted average over the current and previous $L-1$ periods. The weight at lag $l$ is defined as:
\begin{equation}
    w_{t-l} = D^{(l - P)^2}, \quad \text{for each } l \in [0, L)
    \label{eq:adstock-weights}
\end{equation}
where larger values of $D$ produce a more dispersed (longer-lasting) effect, and $P$ controls how many periods the peak effect lags behind the initial exposure. The adstock-transformed series is then:
\begin{equation}
    x_t^* = \mathrm{Adstock}(x_t, \ldots, x_{t-(L-1)};\, L, P, D) = \frac{\sum_{l=0}^{L-1} w_{t-l} \cdot x_{t-l}}{\sum_{l=0}^{L-1} w_{t-l}}
    \label{eq:adstock-transform}
\end{equation}

\paragraph{Bayesian Regression.}
For each channel $c$, we estimate a Bayesian linear regression over $M$ campaigns using PyMC \citep{salvatier2016pymc3}. Let $y_t$ denote the daily channel-level incremental sales contribution (obtained from temporal disaggregation in Stage~2), and let $x_{j,t}^*$ denote the adstock-transformed daily media series for campaign $j \in \{1, \ldots, M\}$ at time $t$. The intercept $\alpha$ captures residual channel-level incremental contribution not explained by the campaign media inputs, and $\beta_j$ denotes the effectiveness coefficient for campaign $j$. Following the Bayesian regression formulation in \citet{bishop2006pattern}, the likelihood is specified as a Normal distribution with mean $\mu_t$ and variance $\sigma^2$, following the notation used in the PyMC documentation \citep{salvatier2016pymc3}:

\noindent\textit{Likelihood:}
\begin{align}
    y_t &\sim \mathcal{N}(\mu_t,\; \sigma^2) \label{eq:likelihood} \\
    \mu_t &= \alpha + \sum_{j=1}^{M} \beta_j \cdot x_{j,t}^* \label{eq:mean}
\end{align}

\noindent\textit{Priors:}
\begin{align}
    \beta_j &\sim \mathrm{TruncatedNormal}(\beta_c,\; \sigma_\beta;\; \text{lower}=0)
    \quad \text{for } j = 1, \ldots, M \label{eq:prior-beta} \\
    \alpha &\sim \mathrm{TruncatedNormal}(\alpha_c,\; \sigma_\alpha;\; \text{lower}=0) \label{eq:prior-alpha} \\
    \sigma &\sim \mathrm{HalfNormal}(10) \label{eq:prior-sigma}
\end{align}
where $\beta_c$ is the aggregate channel-level media effectiveness coefficient derived from MMM, shared as the prior mean across all $M$ campaign-level coefficients, and $\alpha_c$ is the corresponding MMM-derived channel-level offset. The non-negativity constraint on $\alpha$ and $\beta_j$ encodes the economic assumption that baseline contribution and media exposure do not reduce sales, while anchoring the prior means to MMM-derived estimates regularizes the model toward aggregate channel effects. The prior scales $\sigma_\alpha$ and $\sigma_\beta$ are weakly informative hyperparameters set to be sufficiently diffuse to allow the data to dominate posterior inference while preserving the economic constraints. The observation noise prior on $\sigma$ follows a half-normal distribution \citep{salvatier2016pymc3}.

\paragraph{Inference.}
Posterior inference is performed using Automatic Differentiation Variational Inference (ADVI) \citep{kucukelbir2017advi} as implemented in PyMC \citep{salvatier2016pymc3}, chosen for computational scalability when fitting independent models across multiple channels every week.

\section{Experiments \& Results }
IMA was evaluated using three years of historical daily data across media channel. For benchmarking, daily IMA predictions are aggregated to weekly resolution and compared against the corresponding weekly MMM-derived channel contributions, since MMM outputs and benchmark error metrics are defined at the weekly level. Due to privacy constraints, channel names, dates, and absolute revenue magnitudes are not disclosed. 

\subsection{Campaign-Level Multicollinearity}
\begin{figure}[h]
    \centering
    \includegraphics[width=0.6\linewidth]{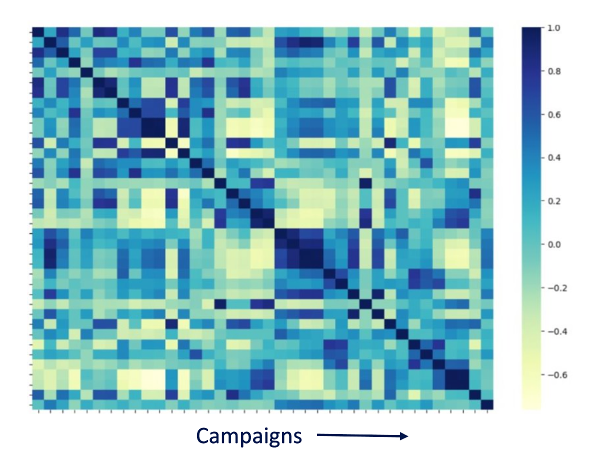}
    \caption{Correlation heatmap of campaigns within a representative channel}
    \label{fig:heatmap}
\end{figure}

We begin by illustrating the structural challenge of campaign-level modeling. Figure ~\ref{fig:heatmap}  presents a correlation heatmap of campaign-level adstock-transformed media variables within a representative channel. Campaigns are ordered alphabetically along both axes, and each cell represents pairwise correlation. 

The heatmap reveals substantial off-diagonal correlations, including visible block structures corresponding to campaign families and overlapping flighting patterns. Such high multicollinearity is typical in large-scale marketing environments. Under standard regression approaches, this degree of correlation would result in unstable coefficients, inflated variance, sign reversals, and overfitting. 

This structural challenge led to the usage of Bayesian regression in IMA. By incorporating MMM-informed priors at the channel level, campaign coefficients are regularized toward economically consistent values, reducing variance while allowing data-driven differentiation where signal strength exists. 

\subsection{Channel-Level Model Fit}

\begin{figure}[h]
    \centering
    \includegraphics[width=0.6\linewidth]{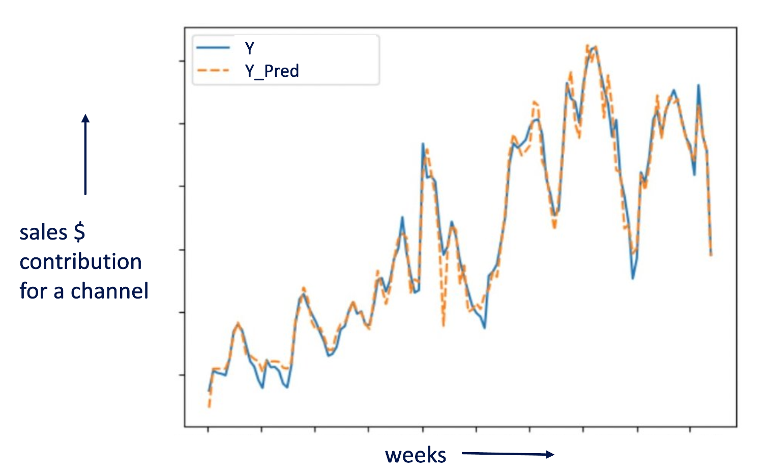}
    \caption{IMA model fit for a representative channel }
    \label{fig:model_fit}
\end{figure}

Although IMA is trained at the daily level, model evaluation is conducted at the weekly level to ensure comparability with MMM benchmarks. Daily predicted incremental contributions are aggregated to weekly totals and compared against the corresponding weekly MMM-derived channel contributions.
Despite the high dimensionality and multicollinearity observed at the campaign level, IMA demonstrates strong consistency with the MMM-derived channel contribution target. Figure~\ref{fig:model_fit} illustrates model fit for a representative channel. The aggregated IMA predictions closely track weekly MMM incremental contributions, achieving an R² of 0.98. This confirms that IMA preserves the aggregate channel signal while operating at campaign-level granularity.
This high fit is notable given that IMA represents a second-stage model, estimating incremental contributions at a lower signal-to-noise ratio than the primary MMM model.

\subsection{Campaign-Level Results}
\begin{figure}[h]
    \centering
    \includegraphics[width=0.6\linewidth]{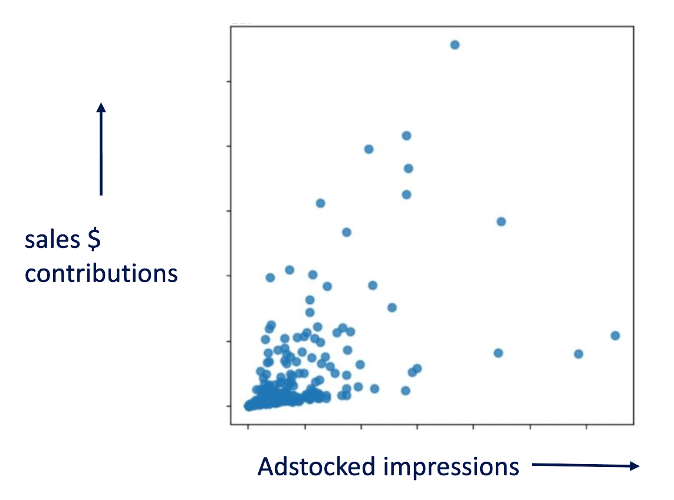}
    \caption{IMA campaign-level results}
    \label{fig:campaign_level}
\end{figure}

Figure~\ref{fig:campaign_level}  illustrates the distribution of campaign-level incremental sales contributions for the same representative channel shown in Figure~\ref{fig:model_fit} over the most recent quarter. Each point represents a campaign, with the x-axis corresponding to aggregated adstock-transformed impressions and the y-axis representing IMA-estimated incremental sales contribution. 

Although campaign coefficients are anchored to a common MMM-derived channel prior, posterior estimates diverge meaningfully based on observed campaign-level signal. The resulting dispersion highlights heterogeneous campaign effectiveness, including low performers, moderate contributors, and high-impact outliers. This demonstrates that Bayesian shrinkage stabilizes estimates without enforcing proportional allocation. Campaign-level contributions are allowed to deviate from the MMM anchor where supported by the data, while remaining consistent with channel-level constraints. 

\subsection{Error Metrics Across Channels}

Using weekly-aggregated predictions for benchmarking, IMA achieves the following Mean Absolute Percentage Error (MAPE) scores across a subset of representative and commonly used channels: 

1. Search: 4.1\%

2. Social: 5.97\%

3. Display: 3.25\%

These error rates are modestly higher than primary MMM benchmark (~2 \%), which is expected for several reasons:

1.	IMA models incremental channel contributions  whereas MMM is optimized on total sales

2.	The incremental contribution targets has lower signal magnitude and higher volatility than total sales

3.	Campaign-level modeling introduces high dimensionality and multicollinearity

4.	IMA operates independently per channel without cross-channel smoothing

Given these structural constraints, sub-6\% weekly MAPE represents strong consistency with MMM-derived channel-level contributions. The results indicate that IMA introduces only a limited degradation in aggregate predictive accuracy while delivering substantial gains in campaign-level granularity.
Overall, the experiments demonstrate that IMA maintains alignment with MMM at the weekly channel level while enabling stable, differentiated attribution at the campaign level within a fully aggregated, privacy-safe framework.

\section{Conclusions}
This paper introduced IMA as a scalable, privacy-safe framework for deriving campaign-level attribution from MMM outputs. IMA addresses the limitations of both user-level attribution and large-scale experimentation need by decomposing MMM-estimated channel contributions into granular campaign-level effects using independent Bayesian models with MMM-anchored priors.

The empirical results demonstrate that IMA maintains strong alignment with MMM at the weekly channel level while enabling differentiated campaign-level attribution. Despite operating at a more granular level, IMA achieves low weekly MAPE (3–6\%) and high reconstruction accuracy (R²). These results indicate that granular attribution can be achieved with only limited degradation in aggregate predictive accuracy. IMA provides stability, interpretability, and scalability without relying on user-level tracking.

IMA is not a conceptual blending of methodologies but a concrete and operational framework validated on three years of real-world marketing data. It is designed to scale across large organizations managing thousands of campaigns concurrently, where experimentation alone may be impractical and user-level attribution infeasible.

Several avenues for future work remain. First, extending IMA to incorporate structured cross-channel interactions could further enhance tactical coordination across media channels. Second, future work may explore dynamic updating of MMM-informed priors through rolling or sequential Bayesian estimation, enabling faster adaptation to changing market conditions and evolving channel effectiveness. Third, integrating experimental results directly into the Bayesian framework as structured prior updates could strengthen causal grounding. Finally, a natural extension of IMA involves incorporating geo-level variation through hierarchical Bayesian modeling. By estimating campaign effects across geographies with partial pooling, IMA could capture regional heterogeneity while preserving statistical stability and improving robustness in sparse data settings. 

IMA represents a step toward a durable measurement paradigm: one that delivers granular campaign-level attribution without relying on deterministic user-level tracking, while preserving consistency with aggregate MMM measurement. By using MMM-derived channel effects as anchors, IMA produces stable campaign-level decompositions that allow meaningful differentiation without estimating each campaign in isolation.

\bibliographystyle{plainnat}
\bibliography{references}

\end{document}